\documentclass[10pt, twocolumn]{article} 
\usepackage[total={6in, 6in}]{geometry}
\usepackage{arxiv}
\usepackage{times}
\usepackage{graphicx}
\usepackage{cite}
\usepackage[ruled]{algorithm2e}
\usepackage{url, hyperref}
\usepackage{amsmath,amssymb,amsfonts}

\usepackage{threeparttable}
\usepackage{multirow}
\usepackage{booktabs}
\usepackage[table,xcdraw]{xcolor}
\usepackage[caption=false, font=footnotesize]{subfig}

\graphicspath{{figure}}

\begin{document}

\title{FV-UPatches: Enhancing Universality in Finger Vein Recognition}

\author{Ziyan Chen$^1$, Jiazhen Liu$^2$, Changwen Cao$^1$, Changlong Jin$^{1}$, and Hakil Kim$^3$ \\
{\normalsize $^1$ School of Mechanical $\&$ Information Engineering, Shandong University} \\
{\normalsize $^2$ School of Information, Renming Universality of China} \\
{\normalsize $^3$ School of Information $\&$ Communication Engineering, INHA University} \\
\\
\today
}

\author{Ziyan Chen, Jiazhen Liu, Changwen Cao, Changlong Jin and Hakil Kim \\
}

\maketitle
\thispagestyle{empty}

\begin{abstract}
\textit{Many deep learning-based models have been introduced in finger vein recognition in recent years. 
These solutions, however, suffer from data dependency and are difficult to achieve model generalization. 
To address this problem, we are inspired by the idea of domain adaptation and propose a universal learning-based framework, which achieves generalization while training with limited data. 
To reduce differences between data distributions, a compressed U-Net is introduced as a domain mapper to map the raw region of interest image onto a target domain. 
The concentrated target domain is a unified feature space for the subsequent matching, in which a local descriptor model SOSNet is employed to embed patches into descriptors measuring the similarity of matching pairs. 
In the proposed framework, the domain mapper is an approximation to a specific extraction function thus the training is only a one-time effort with limited data. 
Moreover, the local descriptor model can be trained to be representative enough based on a public dataset of non-finger-vein images. 
The whole pipeline enables the framework to be well generalized, making it possible to enhance universality and helps to reduce costs of data collection, tuning and retraining. 
The comparable experimental results to state-of-the-art (SOTA) performance in five public datasets prove the effectiveness of the proposed framework.
Furthermore, the framework shows application potential in other vein-based biometric recognition as well.}
% The implementation is available in} \url{https://github.com/celinechanny/FV-UPatches}.
\end{abstract}

\section{Introduction}
Recent years have seen the rapid development of security systems, especially the ones involving biometric traits\cite{review1}.
The popular identification or verification system based on extrinsic biometric modalities, e.g., fingerprint, face, and iris, have exposed much to the public eye. 
However, they are considered vulnerable to malicious attacks such as forgery\cite{acc-ROI}.
While finger vein biometrics, as one of the intrinsic biometric traits, is more desirable for users and has drawn the attention of many researchers due to its excellent advantages: liveness-effective \cite{personal}, immune to counterfeit \cite{review2} and user-friendly \cite{Handbook1}.
Therefore, Finger Vein Recognition(FVR) has been greatly valued in practical application and deployment.

\begin{figure} 
    \centering
    \includegraphics[width=1\linewidth]{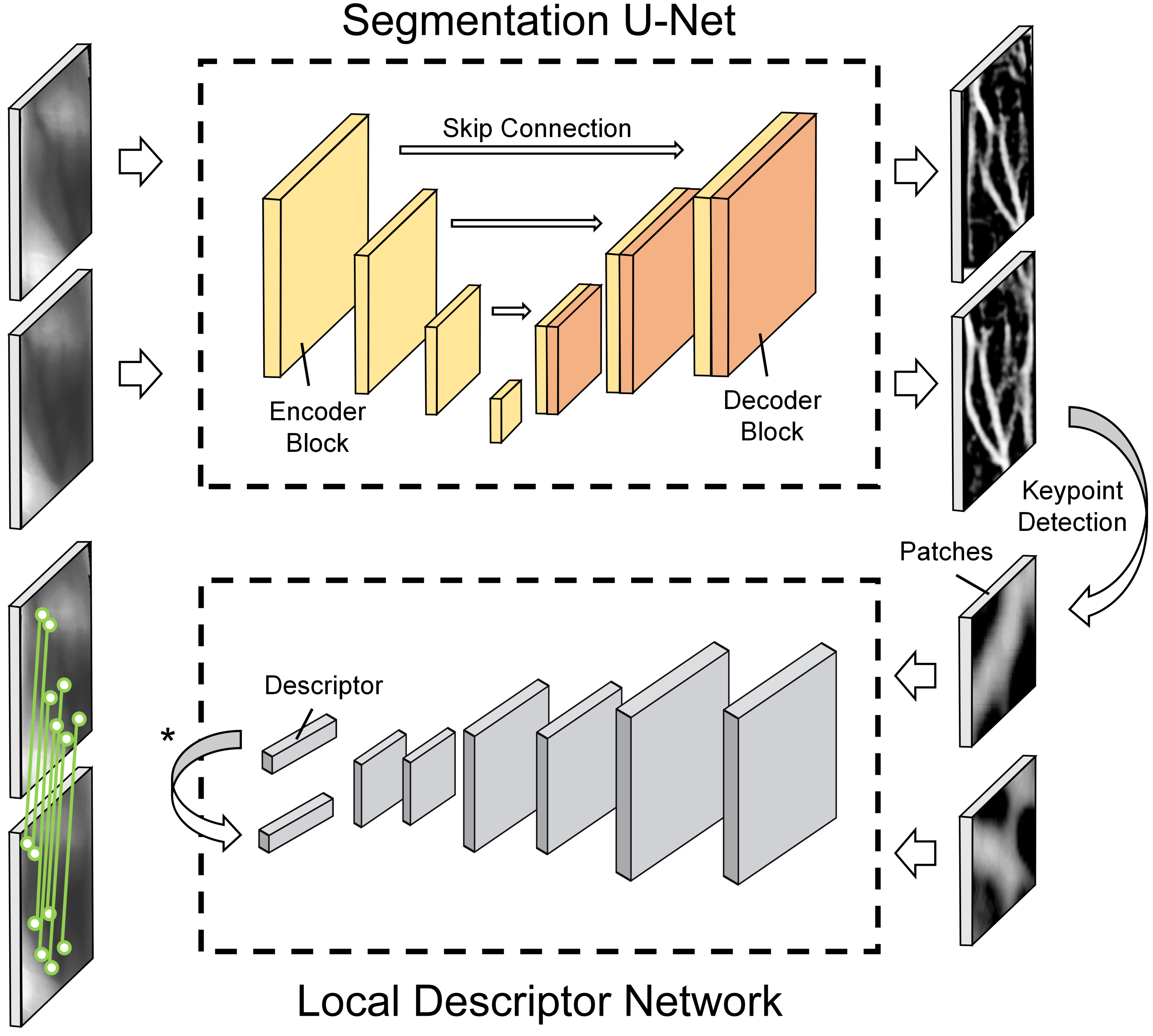}
    \caption{\textbf{Pipeline of the FV-UPatches}. 
        The sign $^{\textbf{*}}$ in the diagram indicates a descriptor matching process.
        The right-hand side of the dotted boxes represents the target domain while the left side represents the raw feature domain.}
  \label{Architecture}
\end{figure}

Recent advances in FVR are mainly focused on recognition challenges posed by the poor image quality and large intra-class distance. 
Among these solutions, many deep learning-based approaches have demonstrated their powerful capability of discriminative feature representation.
For example, FV-Net \cite{FV-Net} leveraged pretrained convolutional neural network to extract regional features of finger veins.
FV-GAN \cite{FV-GAN} employed the powerful Generative Adversarial Network (GAN) to generate vein patterns, and JAFVNet \cite{JAFV-net} introduced joint attention module to further improve the ability of the network to extract discriminative identity features.

The learning-based methods have achieved superior performance when trained and tested in the same dataset. However, most models can only perform well in a single dataset and suffer from poor generalization. It means the tuning or retraining effort is often required when given a different dataset. 
The root cause for this problem lies in the base of traditional machine learning methodologies: assuming that the distribution of training and testing data are from the same domain, such that inputs can be embedded into the same feature space \cite{transfer-review1}.
Away from the assumption, the performance of the learner will be compromised. 
In this work, we propose a universal framework in which different data distributions are transformed to mitigate the data dependency of the model without deviating from the assumption.

As images of the same object could fall into different marginal or conditional distributions when captured in different poses, illuminations, and user states~\cite{transfer1}, the differences in the data distribution of public datasets lead to difficulties in achieving  universality with existing models.
To improve generalization, some researchers had made attempts to leverage fusion training strategies \cite{Siamese}.
These strategies, however, are only to obtain a fused marginal data distribution which is not necessarily indicative of the underlying distribution and can be still biased.

Therefore, it would be natural to come up with an idea: whether it is possible to train a framework that can be universal enough and retraining-free to transfer to a different dataset? 
Many traditional recognition algorithms are data distribution-independent as usually the morphological features extracted in vein pattern maps. 
We leverage the characteristic of traditional methods to help the learners generalize.
Drawing on the idea of domain adaptation in transfer learning \cite{transfer-review1}, the vein pattern extraction can be considered as mapping and concentration of feature distributions.
Based on the work in \cite{U-Net}, we introduce a compressed U-Net as a pattern extraction model, and meanwhile as a domain mapper to help unify different marginal data distributions into a target feature domain.
It allows the subsequent matching network to always behave in the same feature domain; thus, the data dependency can be mitigated.
The hand-crafted morphological features are used to supervise the  training of the U-Net and the approximation to the traditional extraction function allows the network to generalize well.
The model shows a good generalization and can be trained only as a one-time effort.

We then use the image local feature matching to accomplish recognition in the target feature domain since the minutia matching has been proven successful in \cite{SDU-minutia, feature-point}.
A powerful local feature descriptor SOSNet is employed to perform further feature extraction. 
As it is the similarity between features to measure, the model can be trained to be representative enough by simply feeding a public dataset of non-finger-vein images, rather than consuming the already sparse finger vein data.
The main contributions of this work are summarized as follows:
\begin{itemize}
    \item To the best of our knowledge, this work makes the first attempt to enhance universality using a learning-based method in FVR. We propose a universal framework that can reach comparable  performance to SOTA with limited data for training and without any retraining or tuning effort to transfer to other datasets. 
    \item The U-Net is innovatively introduced to achieve domain mapping and probabilistic pattern extraction in FVR. The hand-crafted features are leveraged as soft-label information to supervise the training of the U-Net.
\end{itemize}

The rest of the paper is organized as follows: after a review in recent advances in FVR (section \ref{related-work}), 
we elaborate the proposed FV-UPatches framework (section \ref{proposed-method}) and explore its effectiveness extensively (section \ref{experiments}).
The effects of involved parameters are further discussed in section \ref{discussion} concerning a balance between computation, storage memory, and performance. Then, the conclusion and future work are given in section \ref{conclusion}.

% \begin{figure} 
%     \centering
%     \subfloat[Raw Input\label{Raw-Input}]{%
%       \includegraphics[width=0.43\linewidth]{ori}}
%     \quad \quad
%     \subfloat[Edge Detection\label{Edge Detection}]{%
%         \includegraphics[width=0.43\linewidth]{border}}
    
%     \subfloat[Edge Curve Fitting\label{Edge Curve Fitting}]{%
%         \includegraphics[width=0.43\linewidth]{fit}}
%     \quad \quad
%     \subfloat[Alignment\label{Alignment}]{%
%         \includegraphics[width=0.43\linewidth]{align.pdf}}
%     \caption{\textbf{Alignment in pre-processing.}
%         (a) An example from HKPU-FV \cite{HKPU-FV}. The sample 
%       images will be roughly aligned to extract ROI using edge information.
%       (b) Weighted addition of edge map and the input.
%       (c) Curve fitting for edges and middle-line extraction. 
%       (d) Centerline of the finger will be aligned to coincides with 
%       the middle line of the image.}
%   \label{Preprocess}
% \end{figure}

\section{Related Work} \label{related-work}

The proposed work draws on recent successes of deep neural networks in vein pattern extraction, and the development of local feature matching methods in FVR, which are introduced in detail in this section.

Typically, an FVR system includes stages of image acquisition, pre-processing, feature extraction and matching \cite{review3}. 
The structural characteristics of the finger and uneven illumination in the data acquisition stage have led to reduced image quality.
Many studies employed some strategies to combat the recognition challenges posed by these problems, including vein enhancement and segmentation with traditional methodologies. 
Miura et al. \cite{RLT0} firstly presented a solution for vein segmentation using Repeated Line Tracking (RLT) to address the problem of degraded performance of the minutiae detection algorithm at vein ridges. 
However, the RLT is computationally demanding and tends to introduce line-like trajectory noise that leads to poor results. 
To address these problems,  Song et al. \cite{curve1} extracted fine vein patterns using mean curvature.
Similarly, in the work of Kumar et al. \cite{gabor}, non-linear image enhancement and the use of Gabor filters also allowed the veins to be segmented.

Based on the above methods, Qin and El-Yacoubi \cite{deep-representation} proposed to achieve automatic labeling to avoid costly intensive manual annotation, which is also prone to errors.
These traditional methods have met with some success but tend to be computationally expensive and often suffer from a problem that many thresholds involved need to be manually and iteratively tuned when given a different dataset.
With the development of deep learning, Yang et al. \cite{FV-GAN} leveraged the powerful Generative Adversarial Network (GAN) to generate vein patterns, with segmentation results obtained from traditional algorithms serving as hard labels to supervise the training. 
However, the vein patterns extracted by this framework are coarse-grained, since the based annotation algorithms extract binary segmentation and tend to introduce noise. 
Other methods that achieved fine probabilistic segmentation, i.e. vein-pattern extraction, are usually based on second derivatives of the image, such as the Hessian matrix. Meng et al. \cite{SDU-minutia} calculated the Hessian matrix and a chief curvature map (CCM), to capture the gradient information of the veins effectively by indicating the concave and convex parts of the image. 

In most recognition algorithms, vein-pattern extraction is considered not only a way to reduce sample variation but serves for subsequent feature extraction and matching.
For traditional methods, the hand-crafted features and their combinations are employed for matching. 
Matsuda et al. \cite{feature-point} detected feature points by eigenvalue analysis based on Minimum-Curvature Map (MCM) and combined the vein directions and vein patterns to calculate descriptors.
For learning-based methods, there are generally two main streams according to the recognition ability: closed-set mode and open-set mode\cite{open-set}. 
Models working in closed-set mode can only recognize the identity which has been included in the training set, the algorithm with a CNN classifier are typical examples \cite{close1, close2, close3, close4}.
Though the close-set models are simpler and perform better, they still have difficulties generalizing and have to be retained to recognize a new identity.
Models working in open-set mode tend to involve a metric learning task, i.e. learning to measure similarities between matching pairs instead of predicting the class.
Therefore, such methods can be used in a wider range of scenarios but suffer from performance degradation, which can be reflected in the work of Hu et al. \cite{FV-Net}.
% FVNet: proposed to extract features by dividing the finger vein ROI image into four conceptual regions, and performed region-wise descriptor matching with the help of a VGG-style network. 

In this work, we try to achieve better universality of the FVR based on deep learning methods.
Ronneberger et al. \cite{U-Net} had proposed to use U-Net to extract high-precision biometric patterns. 
Based on this work, we introduce a compressed U-Net as a vein extraction network, and meanwhile as a domain mapper to unify different marginal data distribution into a target probabilistic segmentation domain, which allows the framework work in a much wider range of application scenarios.
% \textbf{Identification and recognition method}. 
\begin{figure}[t]
    \centering
    \subfloat[Input ROI\label{Input-ROI}]{%
       \includegraphics[width=0.45\linewidth]{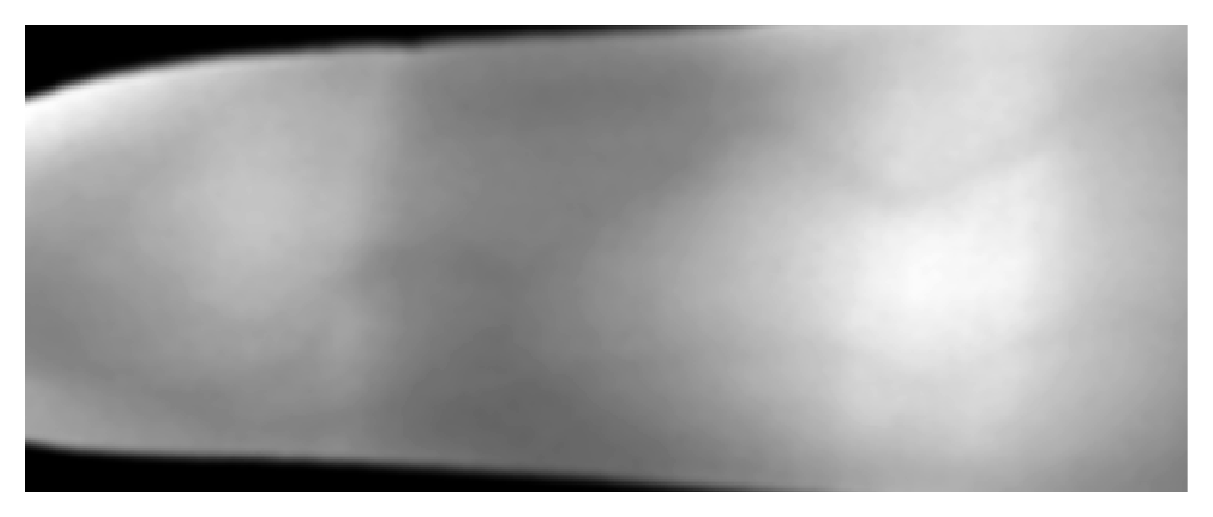}}
    \quad
    \subfloat[Vein pattern\label{Enhanced}]{%
        \includegraphics[width=0.45\linewidth]{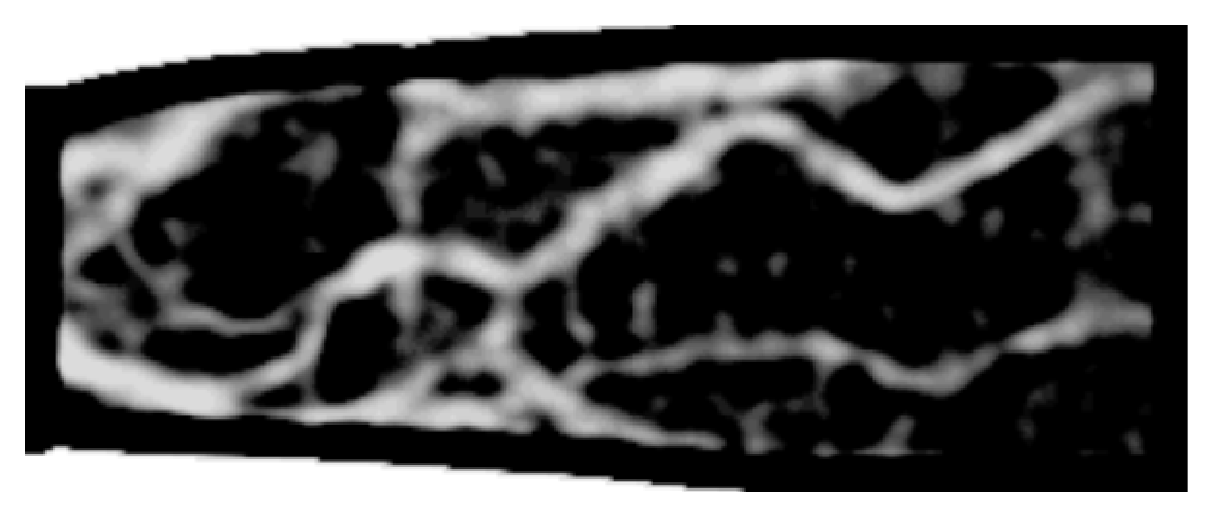}}
    \\
    \subfloat[Gassian filtered\label{Gaussian}]{%
        \includegraphics[width=0.45\linewidth]{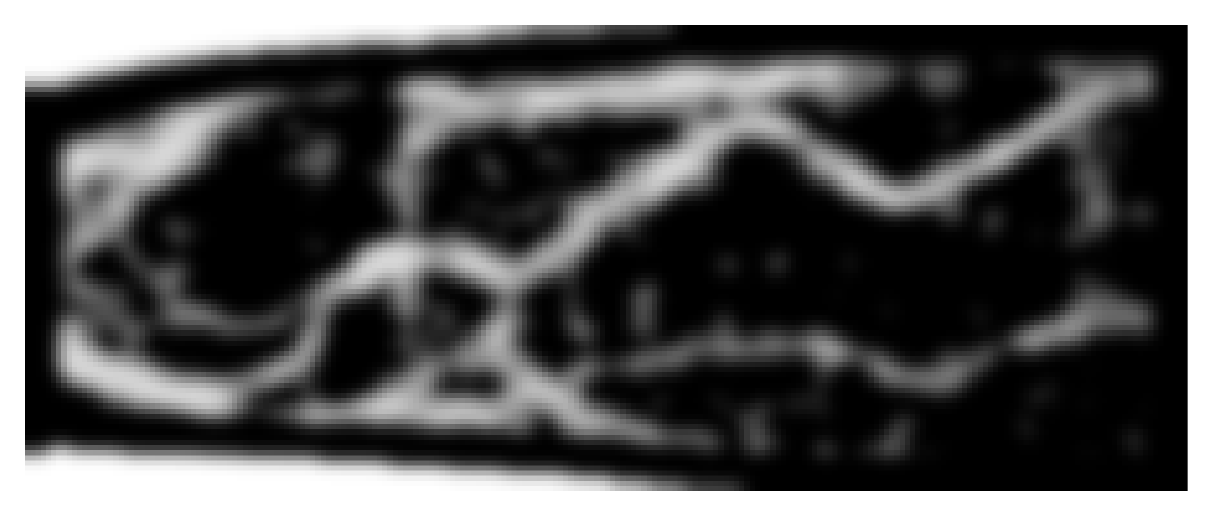}}
    \quad
    \subfloat[Skeleton\label{Skeleton}]{%
        \includegraphics[width=0.45\linewidth]{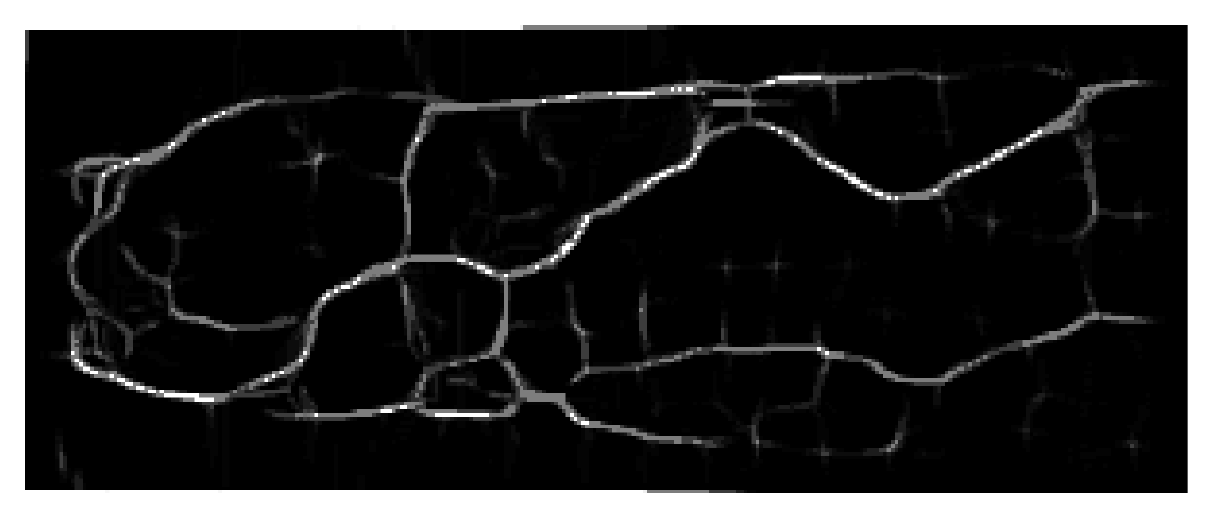}}
    \caption{Vein extraction and ridge keypoint detection.
    (a) ROI of an example from MMCBNU \cite{MMCBNU}.
    The three remaining images demonstrate the pre-processing process.
      (b) is the vein extraction using a traditional second-order central moment method, which is regarded as soft labels of the U-Net.
      (c) (d) are further processed based on (b) to feed to next module.}
  \label{Skeletonisation}
\end{figure}

\section{Proposed Method} \label{proposed-method}
In this section, we \textbf{firstly} focus on preprocessing, in which the region of interest (ROI) of the finger-vein image is aligned and located. 
\textbf{Then} the domain mapping and vein pattern extraction based on a compressed U-Net is introduced. We leverage the hand-crafted features as soft-label information to supervise the training of the U-Net, which mitigates label noise problems caused by automatic annotation. 
\textbf{Finally}, a pre-trained local descriptor model is employed to achieve high-performance minutia matching. 

The proposed framework is named as FV-UPatches, as it is mainly based on a probabilistic U-Net and a patch descriptor model. 
The pipeline of this framework is shown in Figure \ref{Architecture}. 
The models involved can be retraining-free and enhance universality across datasets which are elaborated on in this section and discussed more extensively in section \ref{experiments} and section \ref{discussion} as well.

% \begin{figure}[b]
%     \centering
%     \subfloat[Raw\label{raw-distribution}]{
%         \includegraphics[width=0.32\linewidth]{raw distribution.pdf}
%     }
%     \subfloat[Enhanced\label{enhanced-distribution}]{
%         \includegraphics[width=0.32\linewidth]{enhanced distribution.pdf}
%     }
%     \subfloat[Unified\label{segmented-distribution}]{
%         \includegraphics[width=0.32\linewidth]{segmented distribution.pdf}
%     }
%     \caption{Grayscale probability density distribution from different datasets and different stages. 
%     The vein pattern is mapped into a lighter area to indicate the probability of 
%     vein exists, while the invalid background is mapped into a total dark one.}
%     \label{grayscle-distribution}
% \end{figure}

\subsection{Preprocessing}
The preprocessing in the proposed framework consists only of ROI extraction and Gaussian filtering.
The ROI extraction is to align the centerline of the finger and filter out the background, which facilitates more stable recognition.

Different ROI extraction algorithms are often required due to large differences between datasets.
A dataset with a strong light background like HKPU is well served by the fixed threshold segmentation method to extract its ROI.
Since the method is not general to different datasets, many studies had employed an edge enhancement and extraction-based method.
However, this method can suffer from problems of edge breakage and missing during extraction due to poor image quality such as overexposure or underexposure.

To solve this problem, we designed a more general ROI extraction algorithm. 
% , as is shown in Fig. \ref{Preprocess}
Firstly, a second-order central moment method is used to calculate the pixel-wise orientation intensity and the maximum responses will appear on the finger edge. 
A horizontal Gabor filter is then employed to ensure subsequent fixed-threshold binarization can determine the edges more robustly.
Furthermore, the quadratic curve fitting is used for noise abatement and edge breakage caused by the image quality.
The image is aligned by rotating the centerline of the finger to horizontal, and the ROI is cropped out from the central area of the image.

\begin{figure}[t]
    \centering
    \includegraphics[width=.95\linewidth]{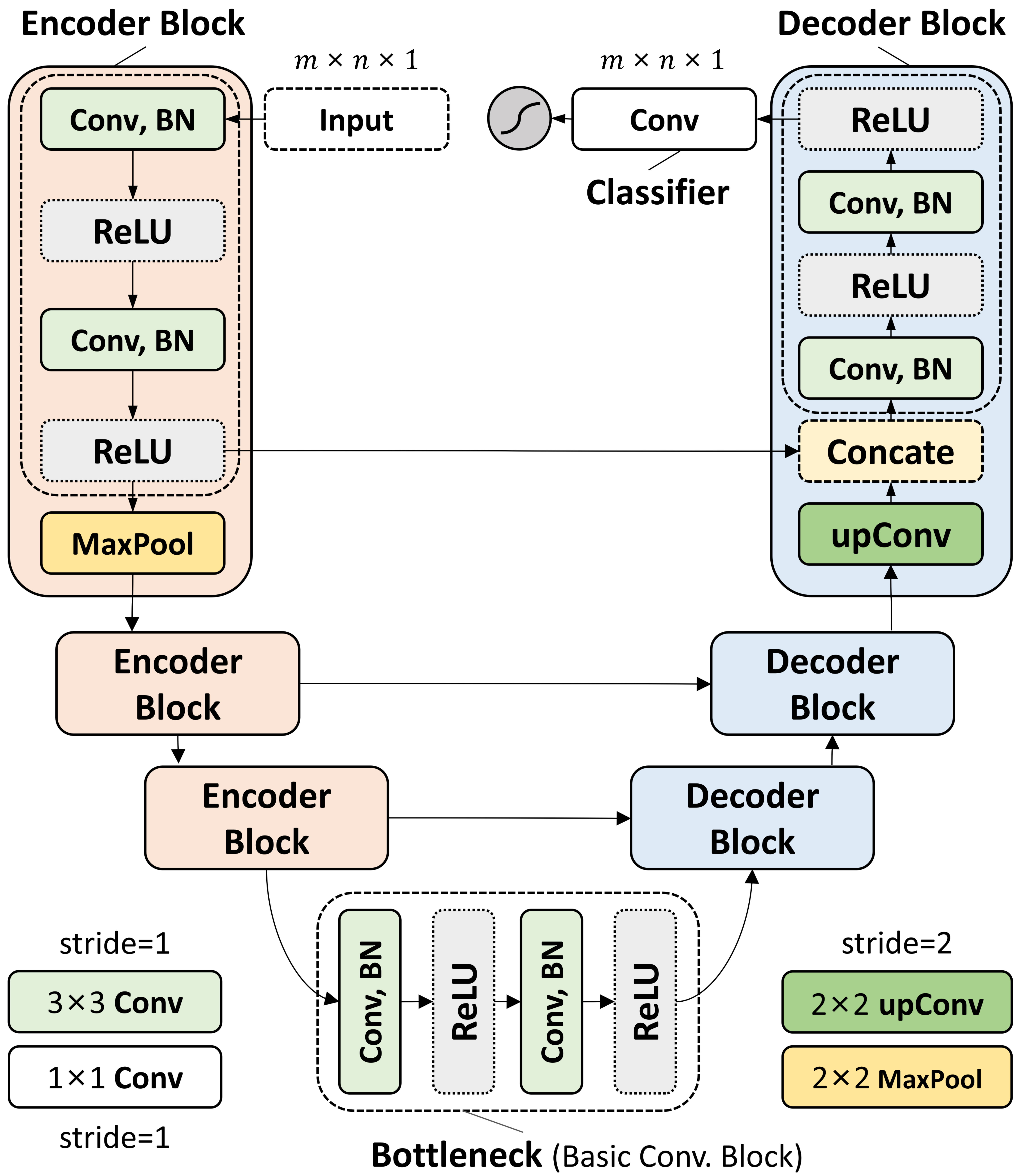}
    \caption{Design of the compressed U-Net architecture. The basic convolutional block selected with black dotted box consists of two $3 \times 3$ convolutions (padding=1) forms the most part of the network.}
    \label{U-Net}
\end{figure}

\subsection{Domain mapping and vein-pattern extraction}
\begin{figure*}[t]
    \centering
    \subfloat[Raw\label{raw}]{%
        \includegraphics[width=0.2\linewidth]{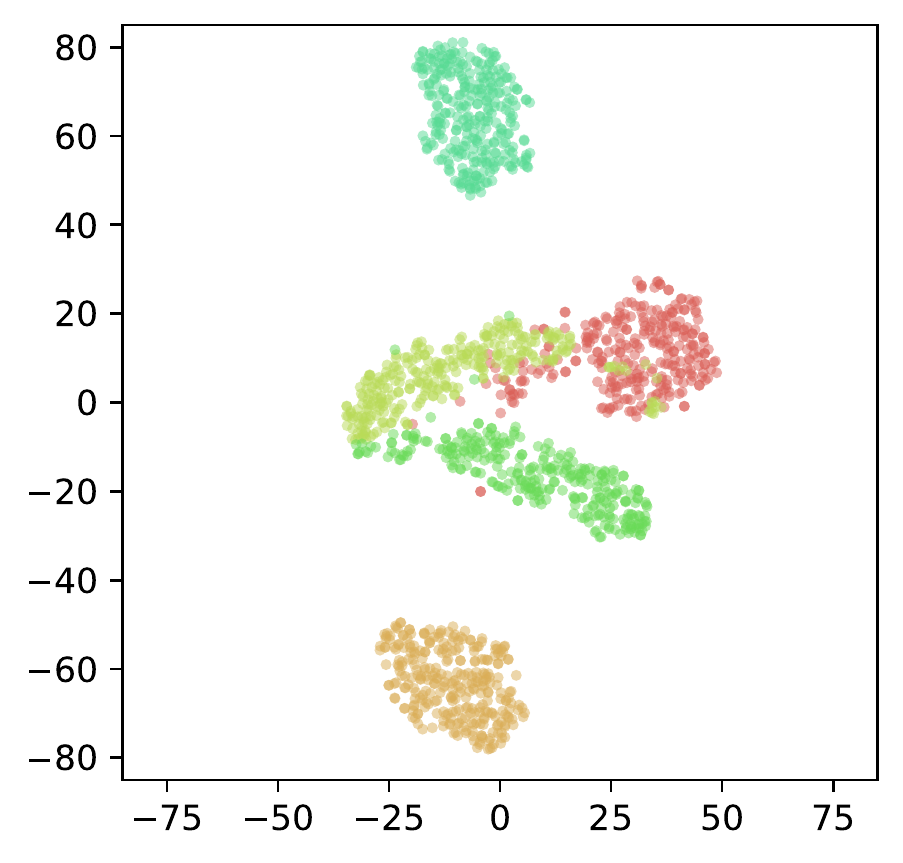}}
    \quad
    \subfloat[Histogram equalized\label{equ}]{%
        \includegraphics[width=0.2\linewidth]{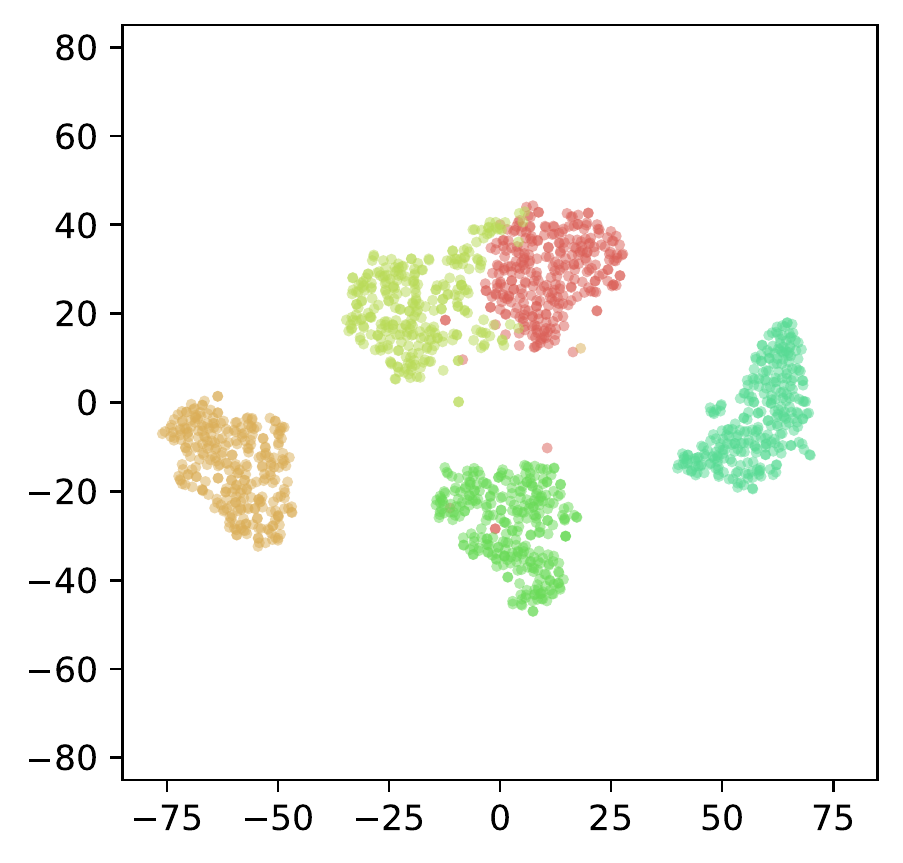}}
    \quad
    \subfloat[Augmented\label{aug}]{%
        \includegraphics[width=0.2\linewidth]{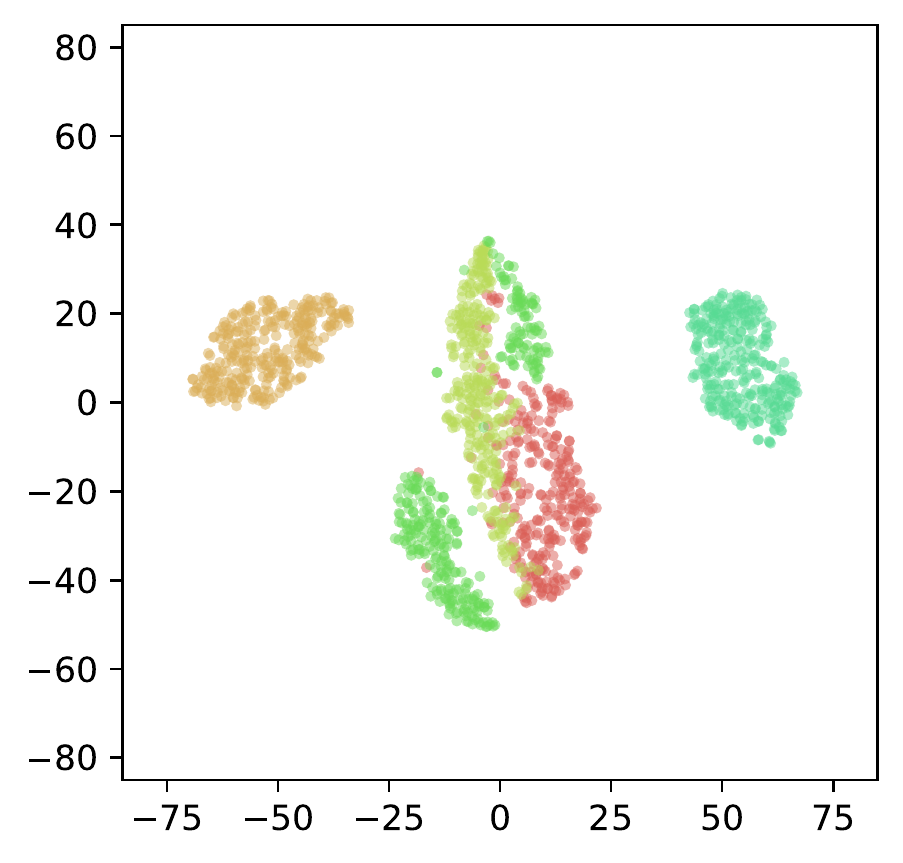}}
    \quad
    \subfloat[Domain mapped\label{seg}]{%
        \includegraphics[width=0.2\linewidth]{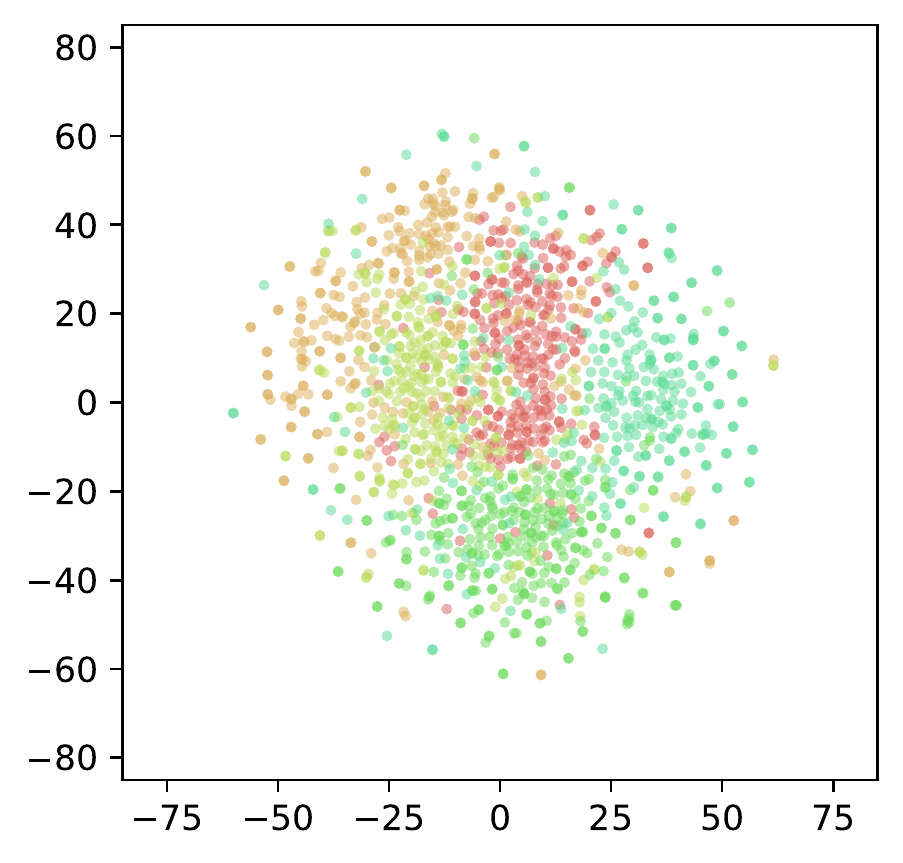}}
    \quad
    
    \caption{Visualization of data distribution using T-SNE on five public datasets. (b) is from data processed with Gamma transformation and CLAHE. (c) is data augmented by random contrast, brightness, Gaussian blur, and saturation. (d) is from the target domain data.}
  \label{T-SNE}
\end{figure*}

Drawing on the idea of domain adaptation in transfer learning, we argue that different data distributions can be mapped into the same target feature domain.
The reduced variation between datasets enables the model to be universal and retraining-free.
The mapping can be achieved by feature transformers with an optimization function.
Based on the characteristics of traditional methods, we consider the vein-pattern extraction itself as an intuitive mapping process to help unify different data distributions.

Most of the current pattern extraction methods are based on traditional algorithms with manual tuning efforts involved when given a different dataset. 
To achieve vein-pattern extraction in a more efficient and fast way, we modify the U-Net \cite{U-Net} to a compressed variant and define the supervised learning as a task of probabilistic segmentation, i.e. a pixel-wise $0$-$1$ classification.

We first obtain the fine vein-pattern map from a traditional method based on second-order central moments. 
The vein-pattern map is further filtered by a Gaussian kernel, which is then taken as soft-label information to supervise the training of the U-Net.
It means that the normalized gray level of each pixel is regarded as a probability that it contains a vein.

This task is usually highly class-unbalanced since the foreground pixels take up only a small fraction.
Thus a combination of \textit{Dice} coeffient \cite{Dice} and \textit{Binary Cross Entropy} (BCE) loss \cite{BCE-loss} is introduced.
Formally, for an input $z \in \mathbb{R}^{m \times n \times 1}$, the expected output probability map denotes as $\mathcal{R}(z) \in \mathbb{R}^{m \times n \times 1}$. 
The optimization problem we solve as:
\begin{equation}
    \begin{gathered}
        \mathcal{L}_{S} = 1 - \frac{2 \cdot 
        {\sum_{i,j}^{m, n} (\hat{y}_{ij} \cdot y_{ij})} + s}
        {\sum_{i, j}^{m, n} (\hat{y}_{ij} + y_{ij}) + s} \\
        % {\xi}_{\mathcal{I}} = {\xi}_{\mathcal{I}}(z, \mathcal{R}(z)) =  \\
        % {\xi}_{\mathcal{U}} = {\xi}_{\mathcal{U}}(z, \mathcal{R}(z)) = 
    \end{gathered}
\end{equation}
where $s$ represents the smoothing factor introduced for further label smoothing.
% The final outputs of the architecture are probability maps of equal size to the inputs.

The design of the modified architecture is illustrated in Figure \ref{U-Net}.
Compared to the original U-Net, the encoder-decoder block with a skip connection is reduced by one layer, and all the channels of the feature map are reduced by a quarter, i.e. $16$, $32$, $64$, $128$ as the channels of the intermediate feature maps at the horizontal encoder-decoder blocks, in which $128$ is the output dimension of the bottleneck.
Even if the model parameters are cut down from $118M$ to $1.86M$ for a lightweight architecture, it is still proven powerful enough to achieve vein pattern extraction properly.
The final $1 \times 1$ convolution and sigmoid layer is to map all the 16-dimensional vectors to the class probability of each pixel.

The pattern extraction itself can be considered as domain mapping.
To illustrate more intuitively, we employ T-SNE for the observation of different data distributions in two-dimensional European space.
The visualization of five public datasets, namely MMCBNU\cite{MMCBNU}, SDUMLA\cite{SDUMLA}, UTFVP\cite{UTFVP}, FVUSM\cite{FVUSM} and HKPU-FV\cite{HKPU-FV}, is shown in Figure \ref{T-SNE}. 
To observe the effect of environmental changes on data distributions, we apply histogram equalization and data augmentation to the original ROI data. 
The clustering results shown in Figure \ref{equ} are from data processed with Gamma transformation and Contrast-Limited Adaptive Histogram Equalization (CLAHE), and the one showed in Figure \ref{aug} is from data augmented by random contrast, brightness, Gaussian blur, and saturation.
It can be discovered a significant difference between datasets enables these data to be clustered properly and sufficiently dispersed.
In contrast, the domain mapped data tend to confuse and fail the clustering.
Figure \ref{seg} shows all feature points are clustered in the central zone as the variation between distributions decreases after domain mapping, which is consistent with the observations of the human eye system. 

Meanwhile, training on the U-Net can be a one-time effort with limited data. Specifically, we select MMCBNU with clearer vein-pattern as the training data and it can generalize well without retraining or fine-tuning effort. 
Since the model is an approximation to the exact traditional extraction function, it can easily transfer to other datasets to extract specific features without retraining or fine-tuning effort. 
Though the compressed network is modified simply without a complicated model-pruning strategy applied, it is still powerful enough to be competent in the task of the specific function approximation, which is validated in section \ref{experiments}.

\subsection{Keypoint detection}
In the last section, we elaborate on the vein-pattern maps that can fall on a unified target domain, which allows the subsequent local descriptor model to behave in the same feature space despite being given a different dataset.

\begin{figure}[b]
    \centering
        \includegraphics[width=0.7\linewidth]{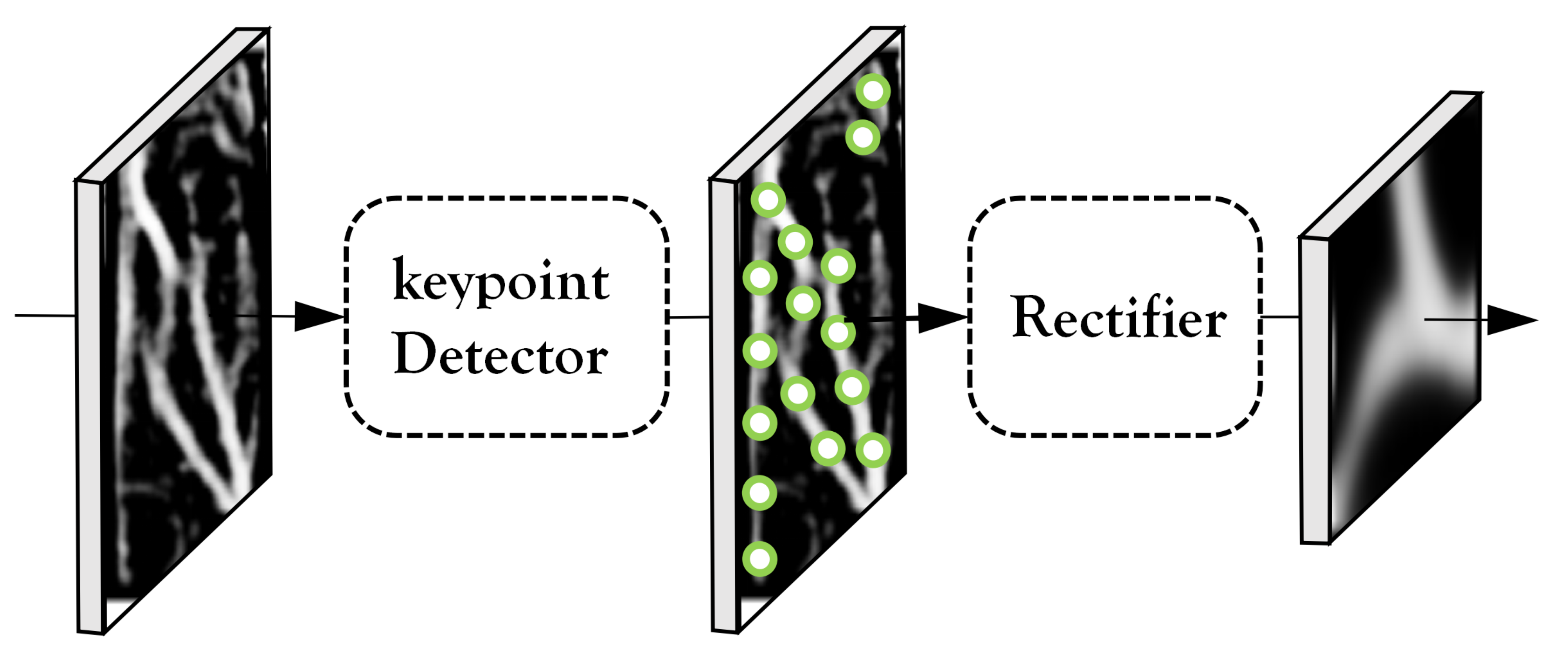}
    \caption{Locate the interest keypoints and rectifing patches around them from the vein pattern map.}
  \label{patchize}
\end{figure}

To achieve descriptor-level matching, the keypoint detection is needed to obtain vein patches as is shown in Figure \ref{patchize}.
The minutia area is cropped as standard patches and will be embedded into descriptors for matching.
Traditional methods used for keypoint detection, such as detector in SIFT \cite{SIFT} and SURF \cite{SURF}, are all proposed as generic detectors for real-world scenarios and typically include keypoint localization, scaling, and orientation assignment. 
It means most of the keypoints detected by these methods are not necessarily located on the vein, where valid identity Information is included.
Therefore, we propose a more straightforward method for keypoint extraction based on the vain-pattern map.

For a vein-pattern map $I$ and a candidate interest point set $p_c(I)$ from the skeleton, the detection process can be elaborated as Algorithm sheet \ref{alg}. 
% and be illustrated in the following three steps: 
% \begin{enumerate}
%     \item Gaussian filtering is applied to the segmentation map to get $\mathcal{G}(I)$, 
%     as shown in Fig.\ref{Gaussian}.
%     \item The result from 1. is skeletonized as $\mathcal{S}(I)$ such as that in Fig. \ref{Skeleton}. The skeleton is regarded as a candidate interest point set $pc(I)$. 
    % \item Determine a suitable parameter of reduction factor $c$ for redundant point removal to get a reduced point set $p(I)$. Since a neighborhood area of the keypoint will be cropped into patches, many of them tend to be redundant to overlap each other. The factor $c$ means the size of a sliding window, and the reduction leaves only the center point and removes other points within the window. It is obvious that the larger the factor $c$ is, the smaller the final point set will be.
% \end{enumerate}
Since the scaling and orientation are trivial in recognition thanks to the alignment of the finger image in preprocessing, we left out the estimation of them for efficient calculations.
It may be counterproductive when these estimations are prone to introduce noise dealing with the linear features.

The proposed method makes the most of the location information of the vein skeleton, enabling most of the patches transformed to contain valid identity information. 

\begin{algorithm}[t]
    \caption{Algorithm for point detection and reduction}\label{alg}
    \KwData{$length(p_c(I)) \geq 20$, $I$, $c$}
    \KwResult{Storing $p(I)$}
    $\mathcal{G}_I \gets GassianFilter(I)$\;
    $\mathcal{S}_I \gets Skeletonize(G_I)$\;
    $pc_I \gets argwhere(S_I > 20)$\;
    $mask \gets Ones(size(I))$\;
    $p_I \gets list\{\varnothing \}$\;
    \For{point $(x, y)$ in $pc$}{
        \If{mask$(x, y)$ is nonzero}{
        $left \gets x-c $\;
        $right \gets x+c $\;
        $up \gets y-c$\;
        $down \gets y+c$\;
        $mask[up:down, left:right] \gets 0$\;
        append $(x, y)$ to $p_I$\;
        }
    }
\end{algorithm}

\subsection{Local feature descriptor}
The patch-based local feature extraction and representation are introduced in this section. 

The methods used to describe local features of images have been widely studied, early traditional methods are mainly based on hand-crafted representations, e.g. gradient distribution, spatial frequency, and differential techniques \cite{des-review}. 
The emergence of annotated patch datasets has led to many data-driven learning descriptor models being proposed, including L2Net \cite{L2Net}, TFeat \cite{TFeat} and HardNet \cite{Hard-Net}, which have achieved encouraging results in image matching. 
In the proposed framework, the L2Net architecture is introduced to embed the patches to $\alpha$-dimention descriptors.

Formally, for the training set with $N$ pairs of patches, the network should embed them into $N$ pairs of descriptors, denoting that the set of descriptors is $\left\{\boldsymbol{x}_{i}\right\}_{i=1 \ldots N}$ and $\{\boldsymbol{x}_{i}^{+} \}_{i=1 \ldots N}$ as its positive correspondence.

The first-order similarity (FOS) loss comes in the form of a Quadratic Hinge Triplet (QHT) loss, enforcing the negatives to be farther away from the positives by a margin:
\begin{equation}
    \begin{gathered}
    \mathcal{L}_{F O S}=\frac{1}{N} \sum_{i=1}^{N} \max \left(0, t+d_{i}^{\mathrm{pos}}-d_{i}^{\mathrm{neg}}\right)^{2} \\
    d_{i}^{\mathrm{pos}}=d\left(\boldsymbol{x}_{i}, \boldsymbol{x}_{i}^{+}\right) \\
    d_{i}^{\mathrm{neg}}=\min _{\forall j, j \neq i}\left(d\left(\boldsymbol{x}_{i}, \boldsymbol{x}_{j}\right), d\left(\boldsymbol{x}_{i}, \boldsymbol{x}_{j}^{+}\right), d\left(\boldsymbol{x}_{i}^{+}, \boldsymbol{x}_{j}\right), d\left(\boldsymbol{x}_{i}^{+}, \boldsymbol{x}_{j}^{+}\right)\right)
    \end{gathered}
\end{equation}
where $d(x_i, x_j)$ is the $L_2$ distance $||x_i-x_j||_2$, $t$ represents the margin. 

Furthermore, the second-order similarity regularization(SOSR) is proposed to further supervise the training:
\begin{equation}
    \begin{gathered}
    \mathcal{R}_{S O S}=\frac{1}{N} \sum_{i=1}^{N} d^{(2)}\left(\boldsymbol{x}_{i}, \boldsymbol{x}_{i}\right)
    \end{gathered}
\end{equation}
The objective function of training goes as follows: 
\begin{equation}
    \mathcal{L}=\mathcal{L}_{F O S}+\mathcal{R}_{S O S}
\end{equation}
More comprehensive information can be found in \cite{SOSNet}.

Training the descriptor model can be based on a dataset of non-finger-vein images since it is the similarity between patches to learn.
The difficult negatives in the public patch-based dataset enable the model to be powerful enough to extract discriminative features between pairs.

\subsection{Matching}
\textbf{Descriptor-level matching}.
We employ L2-distance to measure distance of a descriptor pair and design a two-stage filtering strategy to minimize initial mismatches.
% Formally, for $\{x_i\}_{1 \ldots N}$ represents the descriptor set of the probe, 
% $\{y_i\}_{1 \ldots N}$ indicates that of the sample from the galley. 
% The L2-distance of a descriptor pair are measured as follows:
% \begin{equation}
%     D_{n}(x_i, y_i) = \frac{d(x_i, y_i)}{||y_i||_2} = 
%         \frac{\sqrt{ \sum_{i}^{N}(x_i - y_i)^2 }}{\sqrt{\sum_{i}^{N} y_i^2}}.
% \end{equation}
A threshold parameter $t_d$ is set for the initial filtering of matches with too large a distance, i.e. only distances less than a specific distance will be accepted.
Then we further filter this matching set to ensure that each key point appears only once in the matching set, which is essential for higher accuracy.
Finally, a RANSAC method is employed to estimate the best homography matrix between an image pair based on the descriptor matchings, with only the correct matching pairs to be retained. 

\textbf{Image-lavel matching}.
For a probe and gallery image matching pair, the system needs to make an accept or reject decision comparing the number of descriptor matches with the decision boundary. 
We calculate the decision matches based on the Receiver operating characteristic (ROC) curve in our experiments, and a threshold is used for image matching decisions. 
For an image pair with a descriptor matching number near the decision boundary, the system is prone to make mistakes.
While for consideration of system security, the decision parameter can generally be selected higher manually, making the system rather recognize a bad genuine as an imposter in real application scenarios.

\begin{figure*}[!ht]
    \centering
    \subfloat[MMCBNU\label{MMCBNU-per}]{%
        \includegraphics[width=0.25\linewidth]{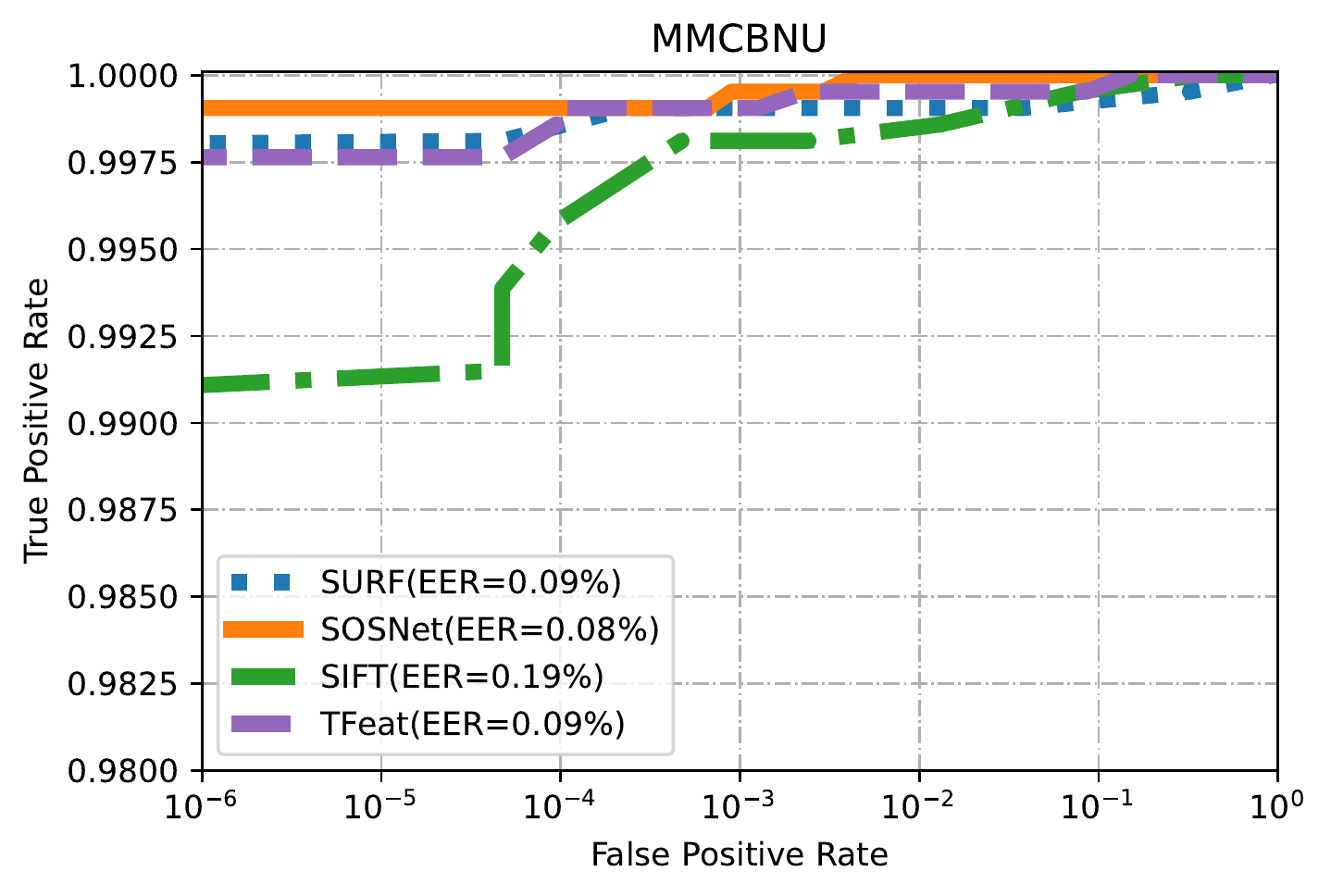}}
    \subfloat[SDUMLA\label{SDUMLA-per}]{%
        \includegraphics[width=0.25\linewidth]{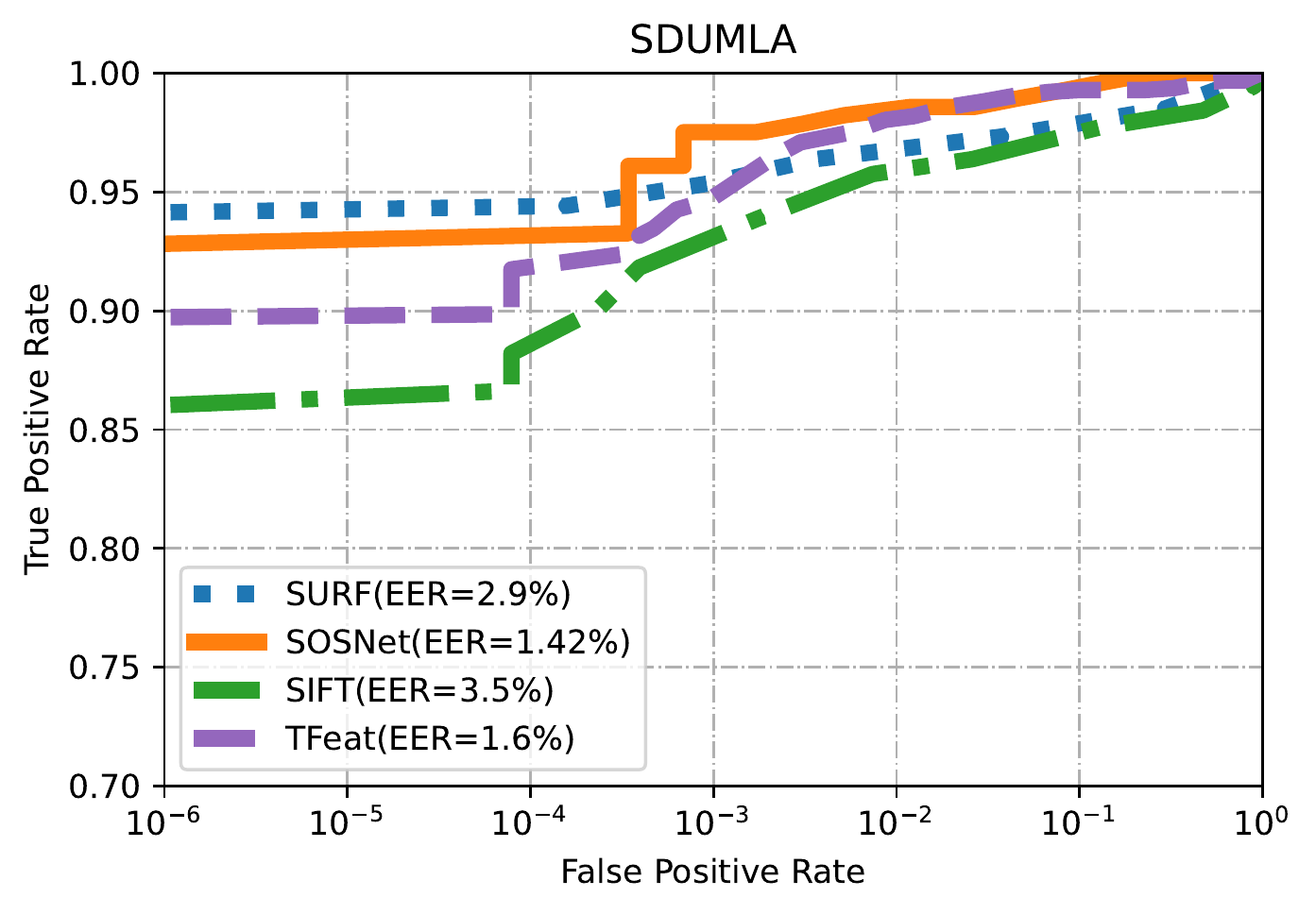}}
    \subfloat[UTFVP\label{UTFVP-per}]{%
        \includegraphics[width=0.25\linewidth]{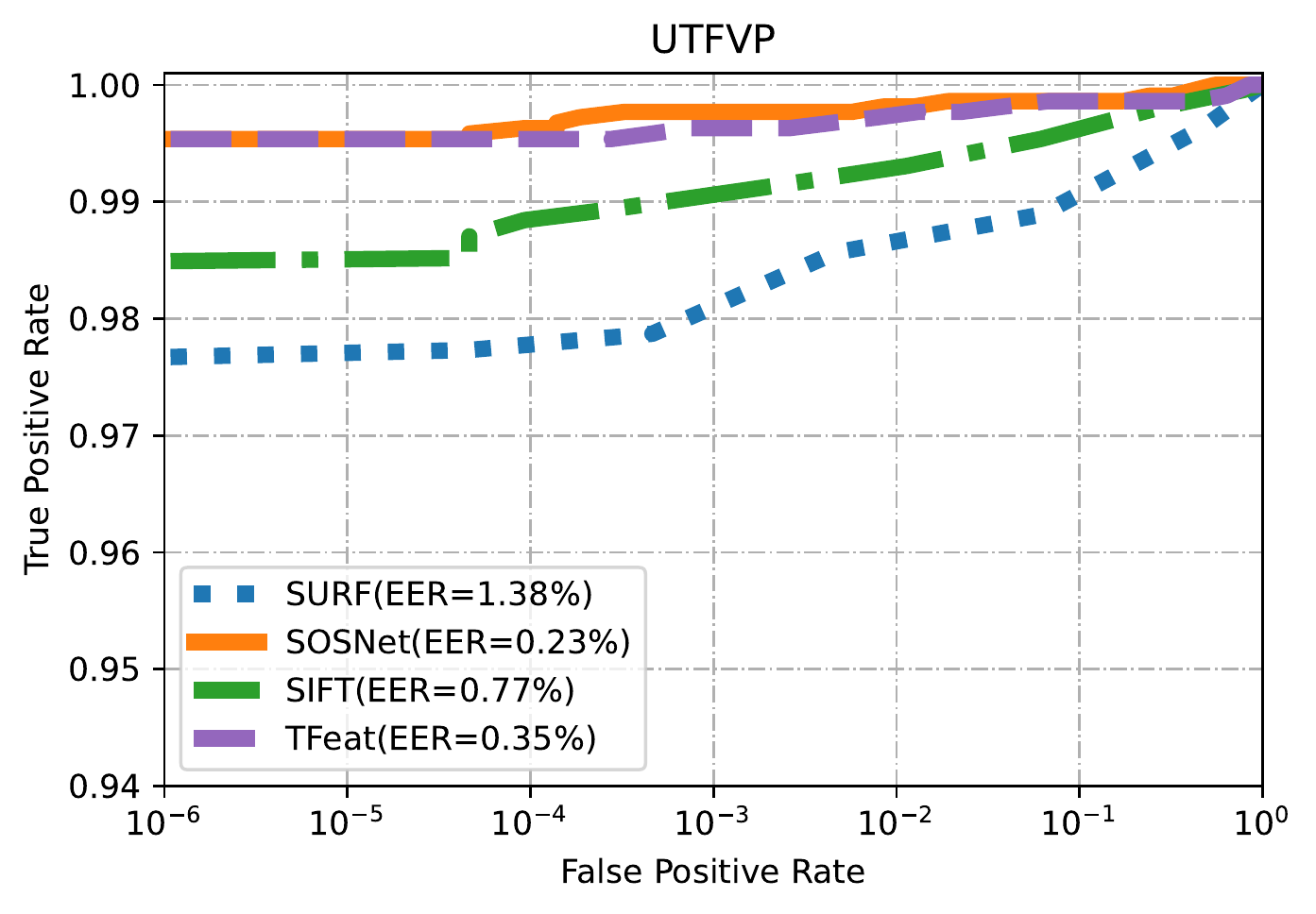}}

    \subfloat[FVUSM\label{FVUSM-per}]{%
        \includegraphics[width=0.25\linewidth]{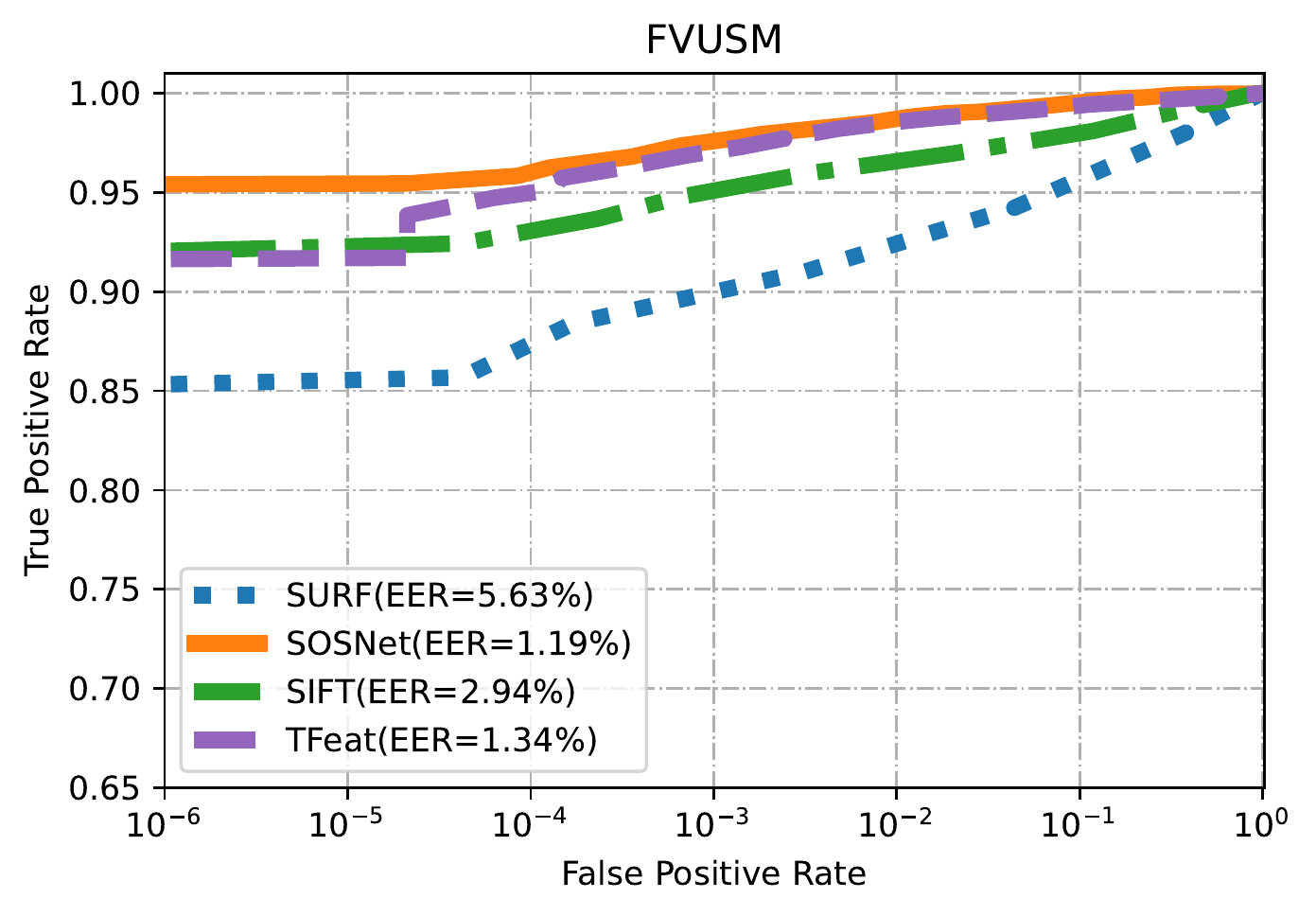}}
    \subfloat[HKPU-FV\label{HKPU-per}]{%
        \includegraphics[width=0.25\linewidth]{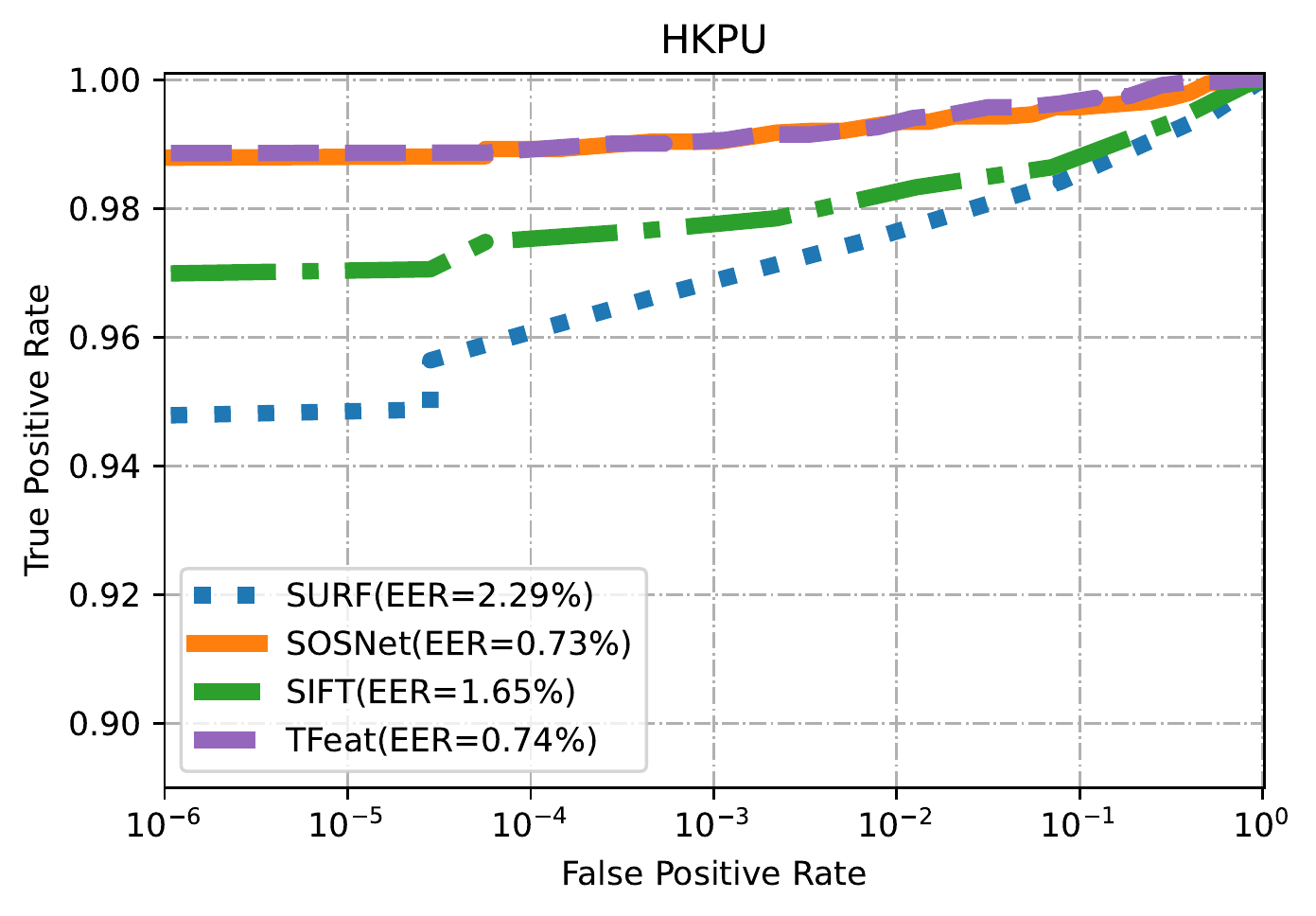}}
    \caption{Performance in five public datasets.}
  \label{performance}
\end{figure*}

\section{Experiments} \label{experiments}
In this section, we evaluate the effectiveness of the proposed FV-UPatches and compare the performance against other SOTA recognition algorithms in five public datasets, whose basic information is introduced in Table \ref{datasets}.

\begin{table}
    \setlength{\abovecaptionskip}{0.5cm}
    \setlength{\belowcaptionskip}{0.1cm}
    \caption{Description to the five public datasets}%%%Table caption goes here
    \label{datasets}
    \centering
    \resizebox{0.9\linewidth}{!}{
    \begin{threeparttable}
    \begin{tabular}{llcc} 
        \toprule
        \textbf{Dataset} & \textbf{Session}   
        & $\#$ \textbf{Sample} & $\#$ \textbf{Class/Subject}  \\ 
        \midrule
        MMCBNU  & Single    & 6,000   & 600/100        \\
        SDUMLA  & Single    & 3,816   & 636/106        \\
        HKPU-FV & Semi-dual* & 3,132   & 312/156        \\
        UTFVP   & Dual    & 1,440   & 360/60         \\
        FVUSM   & Dual      & 5,904   & 492/123        \\
        \bottomrule
    \end{tabular}
    
    \begin{tablenotes}
        \footnotesize
        \item[*] The semi-dual session refers to the fact that only the first 105 subjects in HKPU-FV are dual-session, with the rest only providing singal-session data.
    \end{tablenotes}
  \end{threeparttable}}
\end{table}

\subsection{Evaluation protocol} \label{protocol}
Since the FV-UPatches framework is proposed to be more universal and data-independent, the framework can be evaluated in a way similar to traditional algorithms, i.e. involving all the classes in the dataset instead of using a portion of it as most learning-based methods do. 
We introduce two protocols for mixed-session and cross-session evaluation.

\textbf{Preliminaries}. 
For a dataset consisting of $z$ classes, with $m$ samples in each class and especially $m_sx$ samples in Session-x within a class, the number of genuine and imposter matching pairs are denoted as $N_{g}$ and $N_{i}$, respectively.

\textbf{FVC2004 protocol}.
To validate the performance of different algorithms in a reproducible matching set, we employ the FVC2004 protocol from fingerprint verification contests \cite{FVC2004}. 
For genuine matching, each sample is compared against all remaining samples of the same class. 
While for imposter matching, only the first sample of the class is compared against the first sample of the remaining class. 
The numbers of matching pairs generated with the FVC2004 protocol are:
\begin{equation}
    \begin{gathered}
    N_{g}= C_m^2 \cdot z, \\ %\frac{n_{c} \cdot\left(n_{c}-1\right)}{2}
    N_{i}= C_z^2. %\frac{n_{c} \cdot\left(n_{c}-1\right)}{2}
    \end{gathered}
\end{equation}
The above protocol will be employed for all-session performance evaluation when comparing against other algorithms, and for effectiveness evaluation of the descriptor model.

\textbf{Cross-session FVC protocol}.
We propose a cross-pair protocol for cross-session evaluation with reference to the ISO/IEC 19795-1 suggestion \cite{protocol}.
The Session-1 data is considered as enrolled while the Session-2 is the probe in a matching pair for cross-session evaluation.
For genuine matching, all Session-1 data will combine with all Session-2 data from the same class.
While for imposter matching, only the first Session-1 sample of the class and the first Session-2 sample of another class will form a matching pair. 
The genuine and imposter matching numbers will be:
\begin{equation}
    \begin{gathered}
    N_{g}= m_{s1} \cdot m_{s2} \cdot z, \\ %\frac{n_{c} \cdot\left(n_{c}-1\right)}{2}
    N_{i}= C_z^2. %\frac{n_{c} \cdot\left(n_{c}-1\right)}{2}
    \end{gathered}
\end{equation}
% Specifically, the genuine matching from HKPU, UTFVP, and FVUSM are 7560, 1440, and 17712, respectively, and the imposter matching sets are accordingly 64620, 21945, and 120786.
This protocol is for cross-session evaluations over the effectiveness and robustness of the system. 

\subsection{Implementation details}
% The implementations are introduced according to the order in this section.

The vein extraction network has to be fed with images in the size of $128 \times 256$, 
while the image size in other parts can be arbitrary. 
In our experiments, we uniformly scale the input ROI images to $(90, 225)$ after preprocessing.

The compressed U-Net is trained with Adam \cite{Adam} and the involved hyperparameters are set as $\alpha=0.01$, $\beta_1=0.9$ and $\beta_2=0.999$. 
The smoothing factor $s$ in the loss function is $1$, the learning rate is set to $1e-4$ and the batch size is $16$.
We train the model in a randomly selected dataset and the model can be transferred to other datasets properly.

% The margin of the skeleton images is cleaned to eliminate the effects of finger edges, for actually $10$ pixels being erased at the top and bottom to avoid introducing descriptor matching noise.
For keypoint detection, the orientation and scaling are left out and are fixed to $-1$ and $ks=11$, respectively. 
The point set reduction factor $c$ is fixed as $4$.
Considering the balance between computational speed, storage consumption, and performance, we further discuss the impact of these parameters in section \ref{discussion}.

We use the SOSNet model trained with the patch-based dataset UBC-phototour \cite{UBC-phototour}. 
% It can be transferred directly to finger vein datasets since it is representative enough to learn similarities between local vein matching pairs.
All the patches are rescaled to $32 \times 32$ whatever value $ks$ is selected. 
The batch size is $M=32$ and a descriptor vector dimention is $\alpha=128$.
The threshold $t_d$ is set to $1.2$ for matching.

All the experiments are performed on a private LXD container with a 32-Core Intel Xeon Silver 4110 2.10GHz processor, and NVIDIA GTX 2080Ti GPU of 11GB graphics memory. 

\subsection{Evaluations} \label{evaluations}

\begin{table}[t]
    \caption{Comparison against SOTA methods.}
    \label{SOTA-comparison}
    \centering
    \begin{threeparttable}
    \resizebox{.9\linewidth}{!}{
    \begin{tabular}{@{}crc@{}}
    \toprule
    \multicolumn{1}{l}{Datasets} & \multicolumn{1}{l}{EER(\%)}  & Method                                              \\ \midrule
                                 & \cellcolor[HTML]{FDE9D9}3.37 & \cellcolor[HTML]{FDE9D9}Proposed                    \\
                                 & 2.78                         & \cellcolor[HTML]{EBF1DE}Siamese CNN* \cite{Siamese}               \\
                                 & 2.61                         & \cellcolor[HTML]{DCE6F1}zone-based minutia matching \cite{SDU-minutia} \\
                                 & 2.35                         & \cellcolor[HTML]{EBF1DE}Composite Image \cite{composite}            \\
                                 & 1.59                         & \cellcolor[HTML]{EBF1DE}dual-sliding window \cite{dual-sliding-window}        \\
                                 & 1.2                          & \cellcolor[HTML]{EBF1DE}FV-Net \cite{FV-Net}                     \\
    \multirow{-7}{*}{SDUMLA}     & 1.18                         & \cellcolor[HTML]{EBF1DE}JAFV-Net \cite{JAFV-net}                   \\ \midrule
                                 & 0.63                         & \cellcolor[HTML]{EBF1DE}Siamese CNN \cite{Siamese}               \\
                                 & 0.3                          & \cellcolor[HTML]{EBF1DE}FV-Net \cite{FV-Net}                     \\
                                 & 0.08                         & \cellcolor[HTML]{EBF1DE}JAFV-Net \cite{JAFV-net}                   \\
    \multirow{-5}{*}{MMCBNU}     & \cellcolor[HTML]{FDE9D9}0.05 & \cellcolor[HTML]{FDE9D9}Proposed                    \\ \midrule
                                 & 2.32                         & \cellcolor[HTML]{DCE6F1}dual-sliding window \cite{dual-sliding-window}        \\
                                 & \cellcolor[HTML]{FDE9D9}0.89 & \cellcolor[HTML]{FDE9D9}Proposed                    \\
                                 & 0.76                         & \cellcolor[HTML]{EBF1DE}FV-Net \cite{FV-Net}                     \\
                                 & 0.41                         & \cellcolor[HTML]{EBF1DE}Siamese CNN* \cite{Siamese}               \\
    \multirow{-5}{*}{FVUSM}      & 0.34                         & \cellcolor[HTML]{EBF1DE}JAFV-Net \cite{JAFV-net}                   \\ \midrule
                                 & \cellcolor[HTML]{FDE9D9}0.70 & \cellcolor[HTML]{FDE9D9}Proposed                    \\
                                 & 0.36                         & \cellcolor[HTML]{DCE6F1}zone-based minutia matching \cite{SDU-minutia} \\
    \multirow{-3}{*}{HKPU-FV}    & 0.33                         & \cellcolor[HTML]{EBF1DE}Composite Image \cite{composite}            \\ \midrule
                                 & 0.46                         & \cellcolor[HTML]{DCE6F1}Maximum curvature \cite{cross-database}        \\ 
    \multirow{-2}{*}{UTFVP}      & \cellcolor[HTML]{FDE9D9}0.14 & \cellcolor[HTML]{FDE9D9}Proposed                    \\ \bottomrule
    \end{tabular}}

    \begin{tablenotes}
        \footnotesize
        \item[] Instead of generalising to other datasets, the models of all the listed learning-based methods were trained and tested in the same dataset. 
    \end{tablenotes}
    \end{threeparttable}
\end{table}

In this section, firstly the effectiveness of the system is explored. 
Then the performance of the proposed method is evaluated by comparing it against both the baselines and SOTA methods.
Finally, the cross-session performance of the proposed method is explored as well.

\textbf{Effectiveness of the vein extraction U-Net}.
The modified U-Net shows a significant improvement in training efficiency, for each epoch takes only $19$ seconds and the network converges in about $20$ epochs. 
Training the U-Net is a one-time effort, and the model can be transferred directly to other datasets though trained in only a randomly selected dataset MMCBNU.
The probabilistic maps and the labels are binarily segmented with a threshold of $0.5$ for evaluation of the U-Net.
The precision scores reach $90.82\%$, $89.24\%$, $93.47\%$, $90.82\%$, $88.27\%$ for UTFVP, SDUMLA, MMCBNU, HKPU, and FVUSM datasets, respectively.

\textbf{Effectiveness of the local descriptor model}.
To validate the effectiveness of SOSNet in the proposed system, TFeat \cite{TFeat}, SIFT and SURF are introduced to conduct comparative experiments.
We evaluate them from the perspective of performance, descriptor matching decision boundary numbers (namely decision matches), and total matching distance.
We firstly conduct controlled experiments with SOSNet, TFeat, SIFT, and SURF in the UTFVP.
For SOSNet and TFeat, We simply employ the models pre-trained in the public patch-based dataset.
Figure \ref{performance} shows the ROC curves of the performance in five public datasets, in which the SOSNet shows a powerful similarity metric ability and achieves superior performances.

\textbf{Comparison against SOTA}. 
Methods working in the closed-set mode often perform better but suffer from poor generalization and are difficult to reflect realistic scene performance.
To ensure the validity of the comparison, we compared against only SOTA traditional algorithms and learning-based methods that can work in the open-set mode.
We evaluate the performance of the proposed universal framework in five public datasets.
The experimental results in Table \ref{SOTA-comparison} show it is comparable to the current SOTA methods.
However, it is also worth noting that all the performance of the learning-based methods were tested with only part of each dataset and a large portion has to be divided for model training. 
While the proposed framework is evaluated with all classes in each dataset and requires no re-training efforts when applied to a new dataset.
It means our method is capable of enhancing the universality of the recognition while maintaining excellent performance.
Specifically, our method on SDUMLA underperforms but within an acceptable range and is more plausible from a security point of view.
Many of the samples from SDUMLA tend to be over-rotated, resulting in a significant reduction in the overlap area between a genuine matching and a much larger intra-class distance than that of other datasets.
In this case, the proposed method tends to reject sample pairs with few local matched areas to avoid increasing the security risk, since many of them are hard to discern to be accepted even by human eyes.
% This can also be demonstrated by the superior performance of the proposed system on other datasets with smaller intra-class distances.

\begin{table}[t]
    \caption{Cross-session evaluation results.}
    \label{cross-session-evaluation}
    \centering
    \begin{tabular}{@{}lccc@{}}
    \toprule
    Datasets & FVUSM & HKPU & UTFVP \\ \midrule
    EER(\%)  & 1.41  & 1.10 & 0.14  \\ \bottomrule
    \end{tabular}
\end{table}

\textbf{Cross-session evaluations}.
Although extensive evaluations have been carried out for the above experiments, few algorithms have been evaluated for vein sampling from different sessions. 
The FVUSM, UTFVP, and HKPU datasets provided dual-session finger samples.
Among them, the UTFVP consists of images taken in two distinct sessions in two days. 
The interval between the two sessions of FVUSM is more than two weeks. And the separated session in HKPU is in intervals of one month to six months, with an average interval of 66.8 days.
We employ the cross-session evaluation protocol as is introduced in section \ref{protocol}.
In the cross-session protocol, only the sample from different sessions with the referred one could form a matching pair. 
Compared with the FVC protocol or the fully-connected one, the cross-session evaluations will better reflect the performance of the system in real application scenarios.
The results are listed in Table \ref{cross-session-evaluation}.
Though it shows some performance degradation due to variation between images of different sessions, it still proves the effectiveness and applicability of the proposed system.  

\textbf{Efficiency evaluations}.
Experiments in this part evaluate the operation time efficiency and memory of the models.
The VGG16 \cite{VGG16}, ResNet-152 \cite{ResNet152} and DenseNet-161 \cite{DenseNet161} had been employed to evaluations in \cite{composite}, which is very deep networks with strong power to extract features. 
But they are generally memory-consuming without compression and might be hard to be deployed in some mobile or embedded devices, as is shown in Table \ref{memory}.
The interference time from vein extraction to recognition decision of the proposed system is $129ms$ with $38ms$ for vein pattern extraction and $91ms$ for subsequent recognition when implemented in the Python environment.
We have not compared the inference times directly with those of other algorithms, as differences in experimental computing devices may introduce significant differences in the estimation of running times.

\begin{table}[]
    \caption{Memory estimation of different networks.}
    \label{memory}
    \centering
    \resizebox{\linewidth}{!}{
    \begin{tabular}{lllll}
        \toprule
             & Proposed & VGG16 & ResNet-152 & DenseNet-161 \\ \midrule
    Params(/M) & 1.73     & 132   & 57.5       & 27.7        \\ \bottomrule
    \end{tabular}}
\end{table}

\section{Discussion} \label{discussion}
In this section, we conducted controlled variable experiments to investigate the effect of two parameters, reduction window size $c$ and keypoint scale $ks$, on the performance of the system. 
The impact of these two parameters on the balance between computation, memory, and performance of the system will be also further discussed.

\subsection{Sensitivity analysis}
We explore the balance between performance, storage, and computation by changing some key parameters. 

\begin{figure}[t]
    \centering
    \subfloat[Impact of $c$\label{impact-c}]{%
        \includegraphics[width=0.5\linewidth]{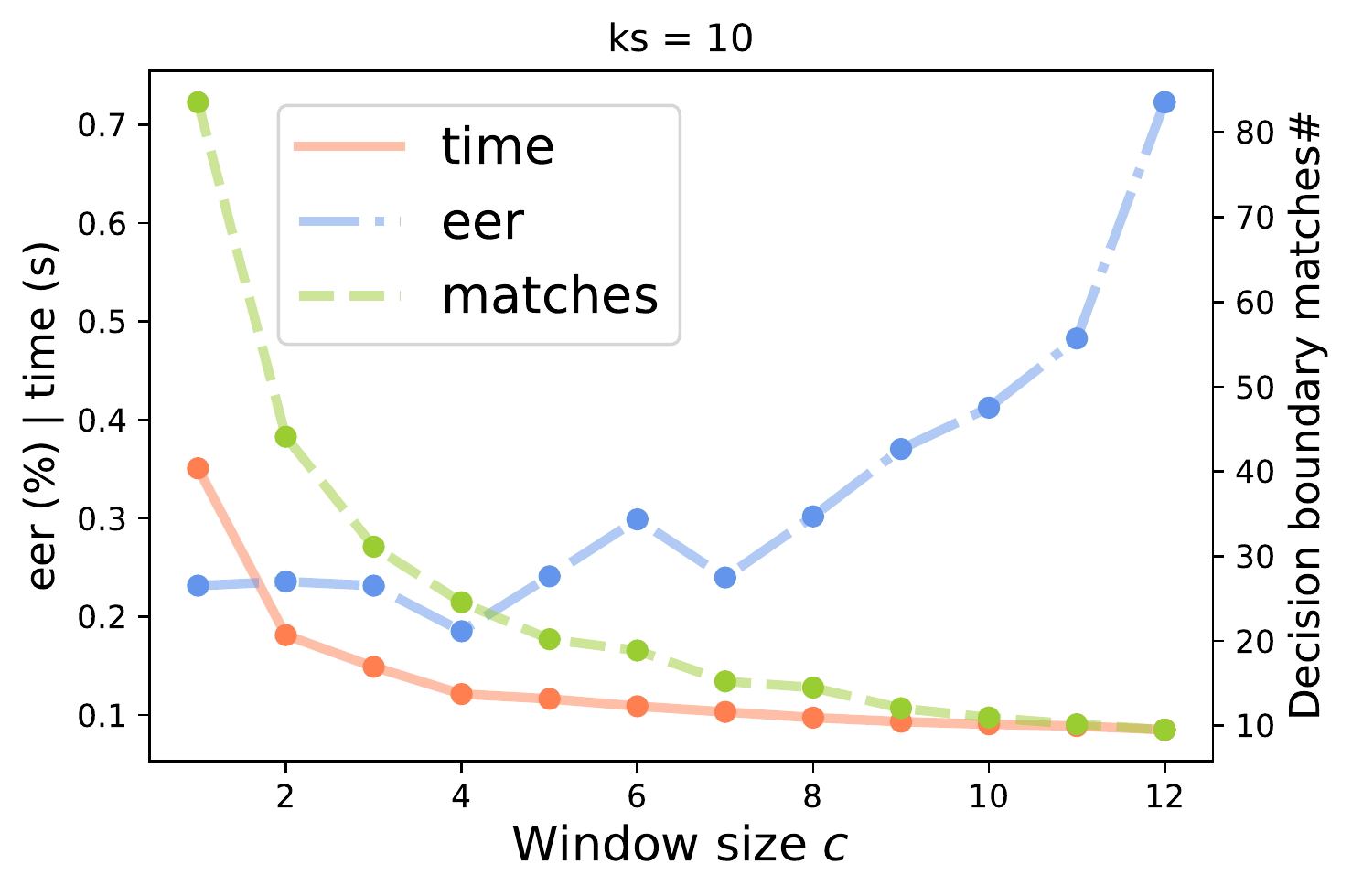}}
    \subfloat[Impact of $ks$\label{impact-ks}]{%
        \includegraphics[width=0.5\linewidth]{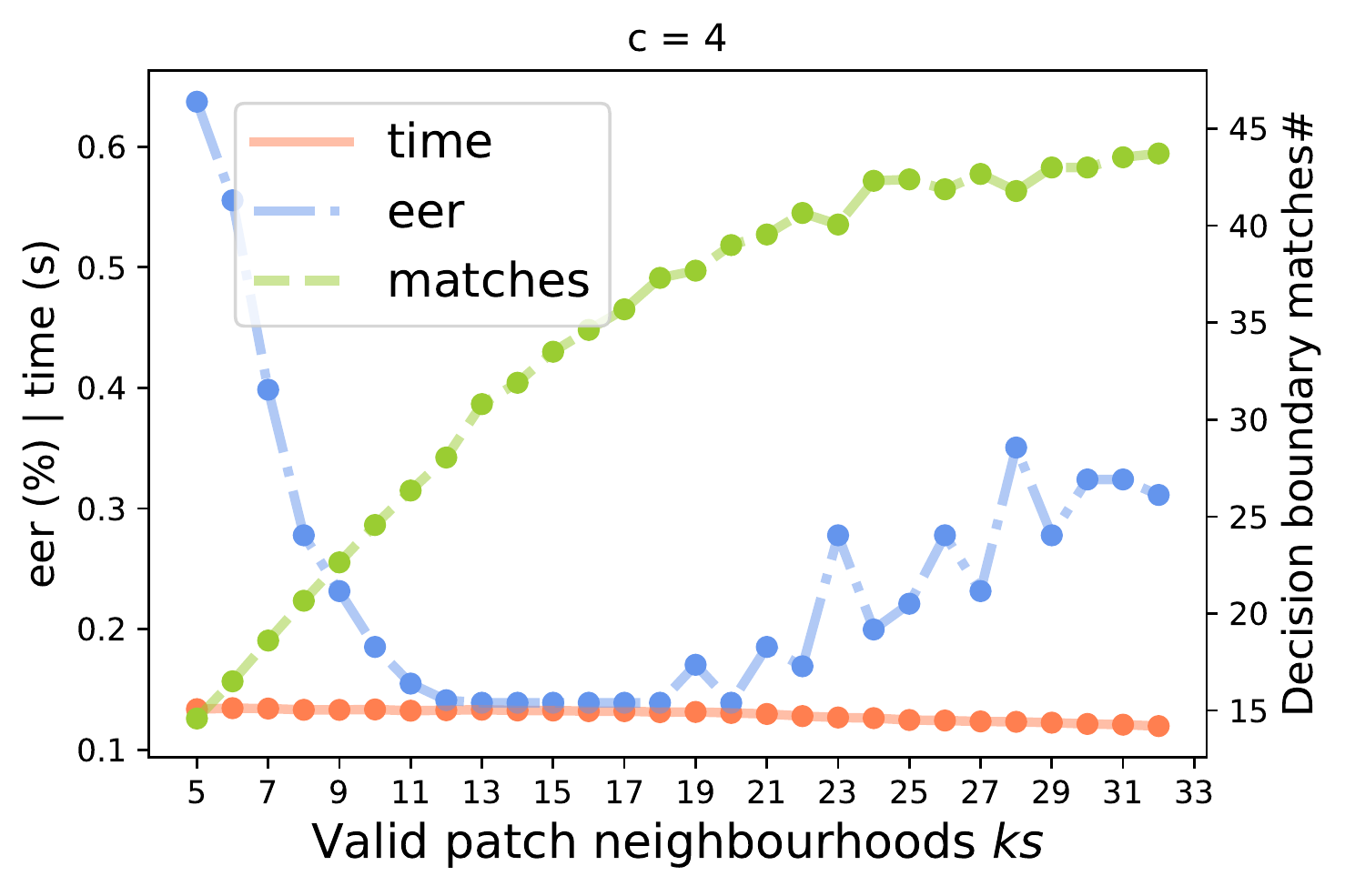}}
    \caption{\textbf{Sensitivity analysis.}
    descriptors have different intervals of distance 
    distribution, we use linear regularisers that do 
    not change the distribution (min-max 
    regularisation) to enable them to be compared.}
  \label{Imapct-c}
\end{figure}

\textbf{Impact of reduction factor $c$}. We keep the keypoint scaling $ks$ fixed to 10, and change the value of the reduction window size $c$ in the interval of $[1, 12]$.
The larger the $c$, the fewer valid descriptors a sample can eventually be extracted, which means that the average system runtime per run will be reduced somewhat and performance may be reduced as is shown in Figure \ref{impact-c}.

\textbf{Impact of keypoint scaling $ks$}. 
To analyze the impact, the reduction factor $c$ is kept fixed to $4$, and $ks$ is changed in the interval $[5, 32]$ in step $1$. 
As is shown in Figure \ref{impact-ks}, the increase of $ks$ does not bring much improvement in system operation efficiency, but the performance and number of matches are affected for more noise may be introduced to the detriment of matching when a larger keypoint scaling is selected.

\section{Conclusion and Future Work} \label{conclusion}
In this work, we proposed a universal framework for flexible and transferable finger vein recognition.
We argued that the vein-pattern extraction can be seen as a mapping process of feature domains, transforming different data distributions into the same target domain.
This allowed us to use a learning-based model based on this target domain without concern about variations between different datasets, thus being  retraining-free and reducing costs of data acquisition, retraining, and tuning.
The experimental results proved the effectiveness of the proposed framework, which shows it can perform properly when fed with a new dataset. 

Future work can focus on more applications in interoperability between devices since the proposed universal framework has the potential to enable recognition across devices.
Moreover, it can be further explored about the universality feature of the proposed framework not only at the cross-dataset level but also at the cross-biometric level. 
The universality of the proposed framework may also lie in its easy applicability to other vein-based biometric recognition.

\end{document}